\documentclass{article}
\usepackage{spconf,amsmath,graphicx}
\usepackage{bm}
\usepackage{url}

\title{DISTILLING THE KNOWLEDGE OF BERT FOR CTC-BASED ASR}
\name{Hayato Futami, Hirofumi Inaguma, Masato Mimura, Shinsuke Sakai, Tatsuya Kawahara}
\address{Graduate School of Informatics, Kyoto University, Sakyo-ku, Kyoto, Japan}

\begin{document}
\ninept
\maketitle

\begin{abstract}
Connectionist temporal classification (CTC) -based models are attractive because of their fast inference in automatic speech recognition (ASR).
Language model (LM) integration approaches such as shallow fusion and rescoring can improve the recognition accuracy of CTC-based ASR by taking advantage of the knowledge in text corpora.
However, they significantly slow down the inference of CTC.
In this study, we propose to distill the knowledge of BERT for CTC-based ASR, extending our previous study for attention-based ASR.
CTC-based ASR learns the knowledge of BERT during training and does not use BERT during testing, which maintains the fast inference of CTC.
Different from attention-based models, CTC-based models make frame-level predictions, so they need to be aligned with token-level predictions of BERT for distillation.
We propose to obtain alignments by calculating the most plausible CTC paths.
Experimental evaluations on the Corpus of Spontaneous Japanese (CSJ) and TED-LIUM2 show that our method improves the performance of CTC-based ASR without the cost of inference speed.
\end{abstract}
\vspace{-10pt}
\section{Introduction}
\vspace{-5pt}
End-to-end automatic speech recognition (ASR) that directly maps acoustic features into text sequences has shown remarkable results.
There are some variants for its modeling: CTC-based models \cite{Graves06-CTC}, attention-based sequence-to-sequence models \cite{Chan16-LAS, Dong18-ST}, and neural network transducers \cite{Graves12-ST, Zhang20-TT}.
Among them, CTC-based models have the advantage of lightweight and fast inference.
They consist of an encoder followed by a compact linear layer only and can predict all tokens in parallel, which is called non-autoregressive generation.
For these advantages, a lot of efforts have continuously been made to improve the ASR performance of CTC-based models \cite{Lee21-ICTC, Nozaki21-RCI}.
In terms of output unit, CTC-based models and transducers are categorized as frame-synchronous models that makes frame-level predictions, while attention-based models are categorized as label-synchronous models that makes token-level predictions.

End-to-end ASR models including CTC-based models are trained on paired speech and transcripts.
On the other hand, much larger amount of text-only data is often available, and the most popular way to leverage it in end-to-end ASR is the integration of external language models (LMs).
In $n$-best rescoring, $n$-best hypotheses obtained from an ASR model are re-scored by an LM, and then the hypothesis of the highest score is selected.
In shallow fusion \cite{Chorowski17-TBD}, the interpolated score of the ASR model and the LM is calculated at each ASR decoding step.
These two LM integration approaches are simple and effective and therefore widely used in the CTC-based ASR.
However, they degrade the fast inference, which is the most important advantage of CTC over other variants of end-to-end ASR.
Specifically, beam search \cite{Graves14-TES} to obtain multiple hypotheses makes CTC lose its non-autoregressive nature.
In addition to beam search, the inference of LM takes much time during testing.

Recently, knowledge distillation \cite{Hinton15-KD} -based LM integration has been proposed \cite{Bai19-LST, Futami20-DKB, Bai21-FESR}.
In this approach, an LM serves as a teacher model, and an attention-based ASR model serves as a student model.
The knowledge of the LM is transferred to the ASR model during ASR training, and the LM is not required during testing.
However, in the formulation of existing studies, the student ASR model has been limited to the attention-based model that makes token-level predictions.
In this study, we propose an extension of this knowledge distillation to the frame-synchronous CTC-based models, so as to integrate the LM while maintaining fast inference of CTC.
We use BERT \cite{Devlin19-BERT} as a teacher LM that predicts each masked word on the basis of both its left and right context.
We have shown that BERT outperforms conventionally-used unidirectional LMs that predicts each word on the basis of only its left context in distillation for attention-based ASR \cite{Futami20-DKB}.
In addition, as recent successful CTC-based models mostly consist of a bidirectional encoder that looks at both left and right context, a bidirectional LM, BERT is suited for a teacher LM.

BERT and attention-based models give token-by-token predictions, while CTC-based models give frame-by-frame predictions.
For distillation from BERT to attention-based ASR \cite{Bai19-LST, Futami20-DKB, Bai21-FESR}, it is obvious that the teacher BERT's prediction for the $i$-th token becomes the soft target for the student attention-based model's one for the $i$-th token.
However, to distill the knowledge of BERT for CTC-based ASR, it is not trivial how to correspond the teacher BERT's token-level predictions to the student CTC-model's frame-level predictions.
In this study, we propose to leverage forced alignment from the CTC forward-backward (or the Viterbi) algorithm \cite{Graves06-CTC} to solve the problem.
During ASR training, the most plausible CTC path attributed to the label sequence is calculated to determine the correspondence between tokens and time frames.
The proposed method improves the performance of CTC-based ASR, even with greedy decoding, without any additional inference steps related to BERT.

\vspace{-5pt}
\section{Preliminaries and related work}
\vspace{-5pt}
\subsection{End-to-end ASR}

\subsubsection{CTC-based ASR}

Let $\bm{X} = (\bm{x}_1, ..., \bm{x}_t, ..., \bm{x}_T)$ denote the acoustic features in an utterance and $\bm{y} = (y_1, ..., y_i, ..., y_L)$ denote the label sequence of tokens corresponding to $\bm{X}$.
An encoder network that consists of RNN, Transformer, or Conformer \cite{Gulati20-CF} transforms $\bm{X}$ into higher-level representations of length $T$.
A CTC-based model predicts CTC path $\bm{\pi} = (\pi_1, ..., \pi_T)$ using the encoded representations.
Let $\mathcal{V}$ denote the vocabulary and $\phi$ denote a blank token.
Then, we define the probability of predicting $v \in \mathcal{V} \cup \{\phi\}$ for the $t$-th time frame as
\begin{align}
P_{\rm CTC}^{(t, v)} = p(v \, | \, \bm{X}).
\end{align}
The output sequence $\bm{y}$ is obtained by $\bm{y} = \mathcal{B}(\bm{\pi})$, where the mapping $\mathcal{B}$ removes blank tokens after removing repeated ones.
The CTC loss function is defined over all possible paths that can be reduced to $\bm{y}$:
\begin{align}
\label{eq:L_CTC}
\mathcal{L}_{\rm CTC} = - \log p(\bm{y} | \bm{X}) = - \sum_{\bm{\pi} \in \mathcal{B}^{-1}(\bm{y})} p(\bm{\pi} | \bm{X}).
\end{align}

\subsubsection{Attention-based ASR}
An attention-based ASR model consists of encoder and decoder networks.
The decoder network predicts each token using the encoded representations and previously decoded tokens.
We define the probability of predicting $v \in \mathcal{V}$ for the $i$-th token as
\begin{align}
P_{\rm Att}^{(i, v)} = p(v \, | \, \bm{X}, \bm{y}_{<i}).
\end{align}
The loss function is defined as the cross-entropy:
\begin{align}
\mathcal{L}_{\rm Att} = - \sum_{i=1}^{L} \sum_{v \in \mathcal{V}} \delta(v, y_i) \log P_{\rm Att}^{(i, v)},
\end{align}
where $\delta(v, y_i)$ becomes $1$ when $v = y_i$, and $0$ otherwise.

\vspace{-5pt}
\subsection{BERT}
BERT \cite{Devlin19-BERT} that consists of Transformer encoders was originally proposed as a pre-training method for downstream NLP tasks such as question answering and language understanding.
BERT is pre-trained on large text corpora for masked language modeling (MLM) objective, where some of the input tokens are masked and the original tokens are predicted given unmasked tokens.
After this pre-training, BERT can serve as an LM that predicts each masked word given both its left and right context.
BERT as an LM has been applied to ASR via $n$-best rescoring \cite{Shin19-ESS, Salazar20-MLMS} and knowledge distillation \cite{Futami20-DKB, Bai21-FESR}.
BERT has been reported to perform better than conventional LMs in ASR thanks to the use of the bidirectional context.

\vspace{-5pt}
\subsection{Distilling the knowledge of BERT for attention-based ASR}
We have proposed to apply BERT to attention-based ASR via knowledge distillation in \cite{Futami20-DKB}.
BERT provides soft labels for attention-based ASR training to encourage more syntactically or semantically likely hypotheses.
To generate better soft labels, context beyond the current utterance is used as input to BERT.
We define BERT's prediction of $v \in \mathcal{V}$ for the $i$-th target as
\begin{align}
\label{eq:P_BERT}
P_{\rm BERT}^{(i, v)} &= p(v \, | \, [\bm{y}^{\rm (pre)}; \bm{y}_{\backslash i}; \bm{y}^{\rm (suc)}]),
\end{align}
where $\bm{y}_{\backslash i}$ is obtained by masking the $i$-th token, that is, $\bm{y}_{\backslash i} = (y_1, ..., y_{i-1}, $\url{[MASK]}$, y_{i+1}, ..., y_L)$.
$\bm{y}_{\backslash i}$ is concatenated with tokens from the preceding utterances $\bm{y}^{\rm (pre)}$ and tokens from the succeeding utterances $\bm{y}^{\rm (suc)}$ to make an input sequence of fixed length $[\bm{y}^{\rm (pre)}; \bm{y}_{\backslash i}; \bm{y}^{\rm (suc)}]$.

The knowledge distillation (KD) loss function is formulated by minimizing KL divergence between $P_{\rm Att}^{(i)}$ and $P_{\rm BERT}^{(i)}$, which is equivalent to minimizing the cross-entropy between them as
\begin{align}
\label{eq:L_KD-att}
\mathcal{L}_{\rm KD} = - \sum_{i=1}^{L} \sum_{v \in \mathcal{V}} P_{\rm BERT}^{(i, v)} \log P_{\rm Att}^{(i, v)}.
\end{align}

The work in \cite{Bai21-FESR} also performed knowledge distillation from BERT to an attention-based non-autoregressive model \cite{Bai20-LASO}, which is a label-synchronous model different from CTC.

\vspace{-5pt}
\subsection{Knowledge distillation for CTC-based ASR}
Knowledge distillation (KD) \cite{Hinton15-KD} between two CTC-based models has been investigated \cite{Andrew15-AM, Kurata19-GCTCP, Takashima18-IKD, Ding19-CCTC}.
The simplest way is to minimize KL divergence between the distributions of the student CTC and those of the teacher CTC frame-by-frame \cite{Andrew15-AM}.
However, it assumes the student and teacher models share the same frame-wise alignment.
This is not true in KD between CTC-based models with different topologies such as KD from bidirectional RNN-based CTC to unidirectional RNN-based CTC \cite{Kurata19-GCTCP, Takashima18-IKD}, which is oriented for streaming ASR applications.
In \cite{Kurata19-GCTCP}, a guiding CTC model encourages student and teacher models to share the same alignment.
Sequence-level KD \cite{Kim16-SKD} was also proposed to address the issue, where $n$-best hypotheses from the teacher CTC are used as targets for the student CTC training \cite{Takashima18-IKD, Ding19-CCTC}.

KD from an attention-based model to a CTC-based model has also been proposed \cite{Moriya20-DAW, Moriya20-SD}.
Token-level predictions from the attention-based model need to be aligned with frame-level predictions from the CTC, which is similar to our KD from BERT to CTC.
Attention weights of $L \times T$ are used for that purpose in \cite{Moriya20-DAW, Moriya20-SD}.
However, for KD from BERT, BERT does not attend acoustic features of length $T$, thus such attention weights cannot be obtained.
Note that KD from an LM including BERT to a CTC-based model is proposed in this study for the first time.

\vspace{-5pt}
\section{Proposed method: Distilling the knowledge of BERT for CTC-based ASR}
\vspace{-5pt}
\label{sec:proposed}

\begin{figure*}[t]
  \centering
  \fbox{
  \includegraphics[width=0.81\linewidth]{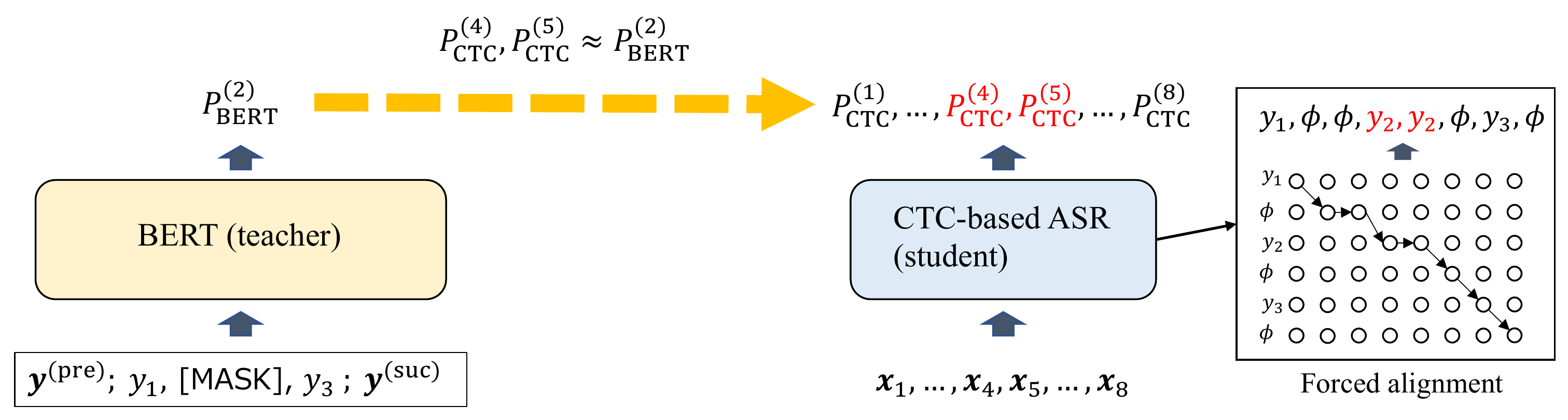}
  }
  \caption{Illustration of our proposed method. The forced alignment path with forward-backward calculation determines which frames $t$ correspond to each token $i$. For $i=2$ ($y_2$), corresponding frames are $t=4$ and $t=5$, so the $4$-th and $5$-th CTC predictions $P_{\rm CTC}^{(4)}$ and $P_{\rm CTC}^{(5)}$ are trained for the $2$-nd soft label from BERT $P_{\rm BERT}^{(2)}$.}
  \label{fig:overview}
  \vspace{-15pt}
\end{figure*}

In this study, we propose to apply BERT to CTC-based ASR via knowledge distillation.
With this method, we expect CTC-based models to further learn the syntactic or semantic relationship between tokens from BERT.
CTC-based models have difficulty in capturing it because they cannot learn it explicitly from the output tokens because of the conditional independence assumption.
Soft labels from BERT help CTC-based models learn it implicitly from acoustic features and intermediate representations in an encoder.
BERT provides token-level soft labels $P_{\rm BERT}^{(i)} (i=1, ..., L)$.
To distill the knowledge of BERT for attention-based ASR that makes token-level predictions $P_{\rm Att}^{(i)} (i=1, ..., L)$, we just minimized KL divergence between $P_{\rm Att}^{(i)}$ and $P_{\rm BERT}^{(i)}$ as in Eq. (\ref{eq:L_KD-att}).
On the other hand, CTC makes frame-level predictions $P_{\rm CTC}^{(t)} (t=1, ..., T)$, so the alignment between $i$ and $t$ is necessary for using token-level soft labels from BERT.

To solve the problem, we propose to use the forced alignment result, that is, the most plausible CTC path $\bm{\hat{\pi}} = (\hat{\pi}_1, ..., \hat{\pi}_T)$ of $\bm{\pi} \in \mathcal{B}^{-1}(\bm{y})$ like \cite{Inaguma21-AKD}, where the CTC path $\bm{\hat{\pi}}$ is used to enable a monotonic chunkwise attention (MoChA) model to learn optimal alignments for streaming ASR.
$\bm{\hat{\pi}}$ can be obtained by tracking the path that has the maximum products of forward and backward variables, which can be obtained in the process of calculating the CTC loss function $\mathcal{L}_{\rm CTC}$ from Eq. (\ref{eq:L_CTC}) with the forward-backward algorithm \cite{Graves06-CTC}.
This method does not introduce any additional architectures or external alignment information (e.g. HMM-based one) for alignment.
For each $i$, $\bm{\hat{\pi}}$ has one or more $\hat{\pi}_t$ that corresponds to $y_i$, and we define the alignment as an one-to-many mapping $\mathcal{A}$ from token index $i$ to frame indices $\bm{t}$.
Note that frame $t$ assigned to $\phi$ in $\bm{\hat{\pi}}$ $(\hat{\pi}_t = \phi)$ does not appear in $\mathcal{A}(i)$.
For example, $\bm{\hat{\pi}} = (y_1, \phi, \phi, y_2, y_2, \phi, y_3, \phi)$ defines $\mathcal{A}(1) = \{1\}$,  $\mathcal{A}(2) = \{4, 5\}$, and $\mathcal{A}(3) = \{7\}$.

Once the alignment $\mathcal{A}$ is obtained, $P_{\rm BERT}^{(i)}$ is used as a soft label for $P_{\rm CTC}^{(t)} $ where $t \in \mathcal{A}(i)$.
For example, given $\mathcal{A}(2) = \{4, 5\}$, the CTC-based ASR is trained to make $P_{\rm CTC}^{(4)}$ and $P_{\rm CTC}^{(5)}$ close to $P_{\rm BERT}^{(2)}$, as illustrated in Fig. \ref{fig:overview}.
The KD loss function is formulated as
\begin{align}
\label{eq:L_KD-ctc}
\mathcal{L}_{\rm KD} = - \frac{1}{\sum_{i=1}^L |\mathcal{A}(i)|} \sum_{i=1}^{L} \sum_{t \in \mathcal{A}(i)} \sum_{v \in \mathcal{V}} P_{\rm BERT}^{(i, v)} \log P_{\rm CTC}^{(t, v)}.
\end{align}

Finally, $\mathcal{L}_{\rm KD}$ is interpolated with $\mathcal{L}_{\rm CTC}$ from Eq. (\ref{eq:L_CTC}) as
\begin{align}
\label{eq:Loss}
\mathcal{L} = (1 - \alpha) \, \mathcal{L}_{\rm CTC} + \alpha \, \mathcal{L}_{\rm KD},
\end{align}
where $\alpha$ $(0 \leq \alpha \leq 1)$ is a tunable hyperparameter.

The alignment is calculated at each training step on the fly.
To mitigate the negative effects of unreliable alignments in early steps, the CTC-based model is pre-trained with Eq. (\ref{eq:L_CTC}).
The soft labels from BERT can be pre-computed for all the training set.
For memory efficiency, top-$K$ distillation \cite{Tan18-MNMT} is applied, where the top-$K$ probabilities of BERT are normalized and smoothed by temperature parameter $T$ to generate soft labels for distillation.
In this study, $K = 8$ and $T = 3.0$ are used.

There is a clear advantage of our method over existing LM integration methods for CTC-based ASR such as $n$-best rescoring and shallow fusion.
Our method introduces the knowledge of LM (BERT) during training, so it has no change in terms of inference time during testing. 
Rescoring and shallow fusion require LM modules during testing, which significantly increases the inference time.
Our method benefits from LM just with greedy decoding without the runtime use of LM, whereas rescoring and shallow fusion requires time-consuming beam search decoding \cite{Graves14-TES} with LM inference.

\vspace{-5pt}
\section{Experimental evaluations}
\vspace{-5pt}
\subsection{Experimental conditions}
We evaluated our method using the Corpus of Spontaneous Japanese (CSJ) \cite{maekawa03-CSJ} and the TED-LIUM2 corpus \cite{Ted214}.
CSJ has two subcorpora of oral presentations, CSJ-APS on academic and CSJ-SPS on general topics.
In CSJ experiments, $240$ hours of transcribed speech of CSJ-APS was used for training ASR.
$7$M-word transcripts of CSJ (both CSJ-APS and CSJ-SPS) and additional $56$M-word text of the Balanced Corpus of Contemporary Written Japanese (BCCWJ) \cite{Maekawa14-BCCWJ} were used for training LMs.
The ASR model and LMs shared the same BPE vocabulary of $10872$ entries.
In TED-LIUM2 experiments, $207$ hours of transcribed speech was used for training ASR, and $250$M-word text in official LM data was used for training LMs.
The BPE vocabulary has $1001$ entries.

CTC-based ASR models consist of Transformer encoder with $12$ layers, $256$ hidden units, and $4$ attention heads.
Adam optimizer with Noam learning rate scheduling \cite{Dong18-ST} of $warmup\_n = 25,000, k = 5$ was used for training the ASR models.
SpecAugment \cite{Park19-SA} was applied to acoustic features, and speed perturbation \cite{Ko15-AA} was also applied in the TED-LIUM2 experiments.
When applying knowledge distillation, CTC-based ASR was pre-trained for $50$ epochs with Eq. (\ref{eq:L_CTC}) and then trained for $50$ epochs with Eq. (\ref{eq:Loss}).
For a fair comparison, baseline CTC-based ASR without knowledge distillation was pre-trained for $50$ epochs and then further trained for $50$ epochs.

We compared three types of LMs: BERT and Transformer LM (TLM) that consist of $6$ layers, $512$ hidden units, and $8$ attention heads, and RNN LM that consists of $4$-layer LSTM with $512$ hidden units.
Adam optimizer of the learning rate of $10^{-4}$ with learning rate warmup over the first $10\%$ of total steps and linear decay was used for training LMs.
During training, sequences of $256$ tokens were fed into LMs, and $8\%$ of tokens in a sequence were masked for BERT.
\vspace{-5pt}
\vspace{-5pt}
\subsection{Experimental results}
Table \ref{tab:csj} shows the ASR results of our proposed method on CSJ.
First of all, CTC-based ASR trained with our proposed knowledge distillation (KD) method \url{(A2)} outperformed a baseline without KD \url{(A1)} in terms of both word error rate (WER) and PPL.
PPL denotes the pseudo perplexity \cite{Chen17-IBRNN} of BERT on the resulting hypotheses:
\begin{align}
    {\rm PPL}(\bm{y}) = \frac{1}{|\bm{y}|}  \sum_{i=1}^{|\bm{y}|} - \log{p(y_i|\bm{y}_{\backslash i})}.
\end{align}
The improvement in both PPL and WER with our method suggests that the knowledge of BERT was indeed incorporated into CTC-based ASR and helped improve WER.
$\alpha$ in Eq. (\ref{eq:Loss}) was determined using the development set, and $\alpha = 0.5$ was found to achieve the best WER for the proposed method on CSJ.
We also observed that increasing $\alpha$ from $0.5$ to $0.7$ improved PPL ($12.87$ to $12.70$) but degraded WER ($9.05$ to $9.12$).
We also trained CTC-based ASR with the recently proposed regularization method, InterCTC \cite{Lee21-ICTC} \url{(B1)}.
Further WER improvement was obtained by training CTC with the combination of InterCTC and our method \url{(B2)}.

Here, we explored a few different KD strategies to find out if there is a better way.
In \url{(A3)}, only the leftmost indices of non-blank tokens in the most plausible path $\bm{\hat{\pi}}$ were used for alignment, while all indices were used in the proposed method.
For example, $\bm{\hat{\pi}} = (y_1, \phi, \phi, y_2, y_2, \phi, y_3, \phi)$ corresponds to $\mathcal{A}(1) = \{1\}$,  $\mathcal{A}(2) = \{4\}$, and $\mathcal{A}(3) = \{7\}$.
In \url{(A4)}, only the rightmost indices were used instead.
In \url{(A5)}, CTC-based ASR was trained with KD from scratch, while KD was applied after pre-training without KD in the proposed method.
In \url{(A6)}, soft labels from TLM were used in KD, that is, $P_{\rm BERT}^{(i, v)}$ in Eq. (\ref{eq:L_KD-ctc}) was replaced with $P_{\rm TLM}^{(i, v)}$.
In \url{(A7)}, one-hot labels were used, that is, $P_{\rm BERT}^{(i, v)}$ in Eq. (\ref{eq:L_KD-ctc}) was replaced with $\delta(v, y_i)$ that becomes $1$ when $v = y_i$, and $0$ otherwise.
This does not use the knowledge of any LM but just encourage CTC's predictions to be aligned with the most plausible path.
Among them, our proposed way described in Section \ref{sec:proposed} performed the best, which demonstrates the effectiveness of our alignment allocation, pre-training, and the use of BERT.

Table \ref{tab:lm-csj} shows the results with other LM integration methods: rescoring and shallow fusion.
Inference times relative to a plain CTC \url{(A1)} measured on CPU are shown in the table as ``InferTime''.
Overall, these methods improved WER more than our KD-based method \url{(A2)}, but they increased inference time far more than the baseline \url{(A1)}.
They took much time for beam search and LM inference.
Note that RNNLM \url{(C3, D3)} can carry over states so is faster than TLM \url{(C4, D4)} in shallow fusion and that TLM \url{(C5, D5)} scores a hypothesis in a single step so is faster than BERT \url{(C6, D6)} in rescoring \cite{Salazar20-MLMS}.
On the other hand, our method did not affect inference time, and WER improvement was obtained with greedy decoding.
Our method also improved the oracle WER \url{(D7)}, and combinations of our method and rescoring or shallow fusion \url{(D3-D6)} further improved WER compared to rescoring or shallow fusion alone \url{(C3-C6)}.

\begingroup
\renewcommand{\arraystretch}{1.1}
\begin{table}[t]
  \caption{ASR results on CSJ with proposed knowledge distillation (KD) -based LM integration. Different KD strategies are compared.}
  \vspace{10pt}
  \label{tab:csj}
  \centering
  \begin{tabular}{lcc}  \hline
     & {WER(\%) $\downarrow$} & PPL (BERT) $\downarrow$ \\
     & eval1 & eval1 \\ \hline
    \url{(A1)}CTC & $9.34$ & $13.70$ \\
    \url{(A2)}+KD (BERT) & $\bm{9.05}$ & $\bm{12.87}$ \\
    \url{(A3)}+KD (BERT) {\small (leftmost)} & $9.15$ & $13.28$ \\
    \url{(A4)}+KD (BERT) {\small (rightmost)} & $9.14$ & $13.19$ \\
    \url{(A5)}+KD (BERT) {\small (scratch)} & $9.59$ & $14.03$ \\
    \url{(A6)}+KD (TLM) & $9.23$ & $13.25$ \\
    \url{(A7)}+Alignment & $9.32$ & $13.79$ \\ \hline
    \url{(B1)}InterCTC \cite{Lee21-ICTC} & $9.22$ & $13.66$ \\
    \url{(B2)}+KD (BERT) & $\bm{8.89}$ & $\bm{12.77}$ \\ \hline
 \end{tabular}
 \vspace{-10pt}
\end{table}
\endgroup

\begingroup
\renewcommand{\arraystretch}{1.1}
\begin{table}[t]
  \caption{Comparison and combinations with rescoring (Resc) and shallow fusion (SF). ``BS'' means beam search without LM, and $b$ denotes beam width. $n$ denotes the number of hypotheses to rescore.}
  \vspace{10pt}
  \label{tab:lm-csj}
  \centering
  \begin{tabular}{lcc}  \hline
     & {WER(\%) $\downarrow$} & InferTime $\downarrow$ \\
     & eval1 & eval1 \\ \hline
    \url{(A1)}CTC & $9.34$ & $\times 1.0$ \\
    \url{(C2)}+BS ({\small $b=5$}) & $9.29$ & $\times 2.2$ \\
    \url{(C3)}+SF (RNNLM, {\small $b=5$}) & $8.95$ & $\times 20$ \\
    \url{(C4)}+SF (TLM, {\small $b=5$}) & $8.71$ & $\times 120$ \\
    \url{(C5)}+Resc (TLM, {\small $n=5$ / $50$}) & $8.89$ / $8.61$ & $\times 3.1$ / $\times 600$ \\
    \url{(C6)}+Resc (BERT, {\small $n=5$ / $50$}) & $8.78$ / $\bm{8.49}$ & $\times 32$ / $\times 890$ \\
    \url{(C7)}Oracle ({\small $n=5$ / $50$}) & $7.25$ / $5.45$ & - \\ \hline
    \url{(A2)}CTC+KD (BERT) & $9.05$ & $\times 1.0$ \\
    \url{(D2)}+BS ({\small $b=5$}) & $9.01$ & $\times 2.2$ \\
    \url{(D3)}+SF (RNNLM, {\small $b=5$}) & $8.65$ & $\times 20$ \\
    \url{(D4)}+SF (TLM, {\small $b=5$}) & $8.59$ & $\times 120$ \\
    \url{(D5)}+Resc (TLM, {\small $n=5$ / $50$}) & $8.61$ / $8.45$ & $\times 3.1$ / $\times 600$ \\
    \url{(D6)}+Resc (BERT, {\small $n=5$ / $50$}) & $8.40$ / $\bm{8.20}$ & $\times 32$ / $\times 890$ \\
    \url{(D7)}Oracle ($n=5$ / $50$) & $7.03$ / $5.36$ & - \\ \hline
 \end{tabular}
 \vspace{-10pt}
\end{table}
\endgroup

\begingroup
\renewcommand{\arraystretch}{1.1}
\begin{table}[t]
  \caption{ASR results on TED-LIUM2.}
  \vspace{10pt}
  \label{tab:ted2}
  \centering
  \begin{tabular}{lccc}  \hline
     & \multicolumn{2}{c}{WER(\%) $\downarrow$} & PPL (BERT) $\downarrow$ \\
     & test & dev & test \\ \hline
    CTC & $12.13$ & $12.57$ & $5.53$ \\
    CTC+KD (BERT) & $\bm{11.40}$ & $\bm{11.79}$ & $\bm{4.92}$ \\
    CTC+KD (TLM) & $11.72$ & $12.28$ & $5.15$ \\ \hline
 \end{tabular}
\end{table}
\endgroup

\begingroup
\begin{figure}[t]
  \centering
  \includegraphics[width=0.95\linewidth]{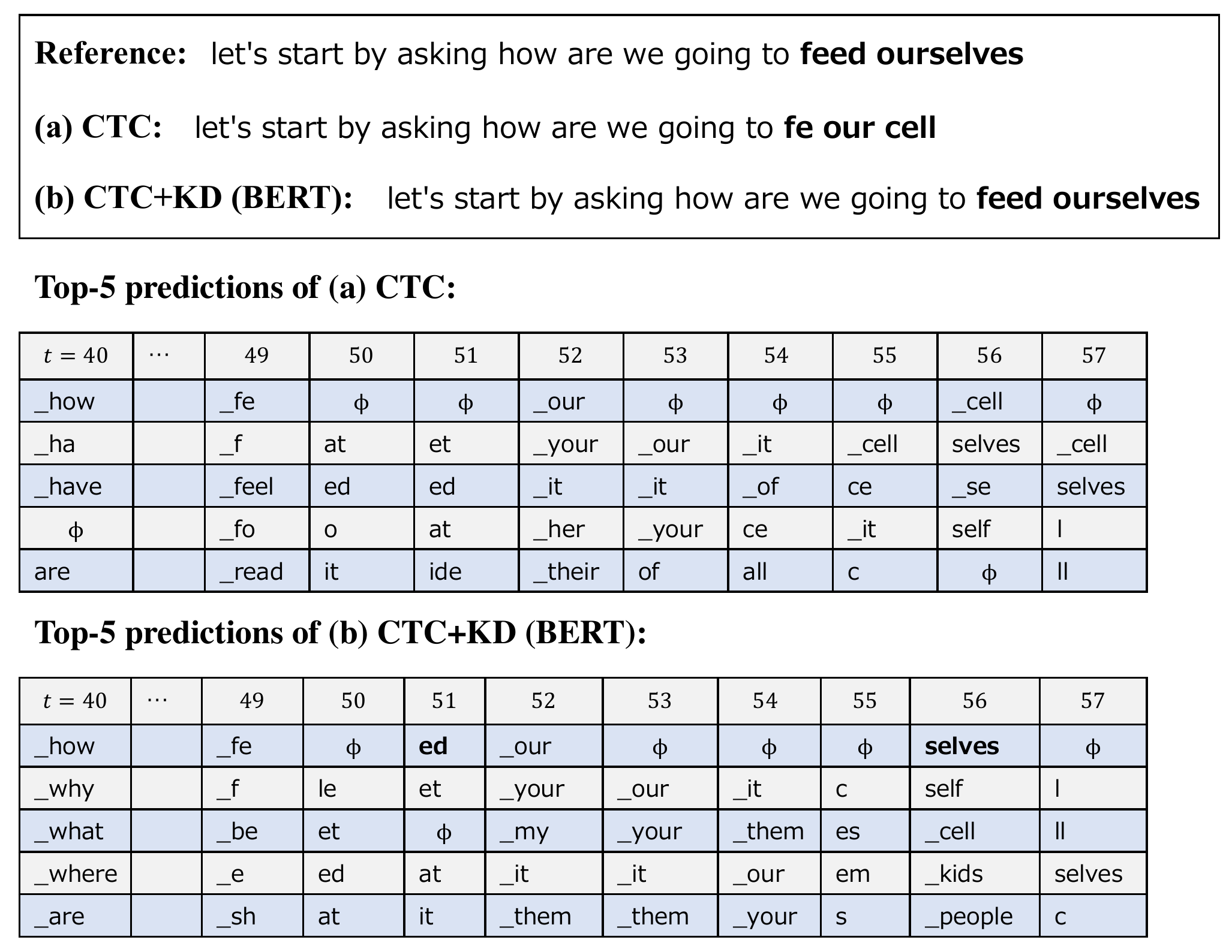}
  \caption{An example of decoded hypotheses and top-$5$ predictions (the upper row is more probable) from two CTC-based models on TED-LIUM2. ``\_'' denotes a word boundary.}
  \label{fig:example}
\end{figure}
\endgroup

Table \ref{tab:ted2} summarizes ASR results on TED-LIUM2, and it shows that our proposed KD with BERT improved WER and PPL of CTC-based ASR as well.
Fig. \ref{fig:example} shows a decoding example for an utterance from TED-LIUM2.
The box at the top of the figure shows decoded hypotheses from two CTC-based models, (a) one trained without KD and (b) the other trained with our method.
While ``feed ourselves'' was erroneously recognized as ``fe our cell'' without KD, it was correctly recognized with KD.
The lower part of the figure shows the top-$5$ frame-level predictions from the two models for the utterance.
We see that the probabilities of semantically plausible subwords ``ed'' and ``selves'' become higher with KD, leading to correct recognition.
It is also interesting to see that semantically plausible subwords such as ``why'' ($t=40$) and ``\_kids'' ($t=56$) appear in the top-$5$ predictions with KD, which indicates the model considers the relationship between tokens in the context much more than the baseline.

\vspace{-5pt}
\section{Conclusions}
\vspace{-5pt}
In this study, we have proposed knowledge distillation-based BERT integration for CTC-based ASR.
For knowledge distillation, BERT provides token-level soft labels, while CTC-based ASR makes frame-level predictions.
We obtained the alignment between them by calculating the most plausible CTC paths.
Our method does not add any computational costs during testing, which maintains the fast inference of CTC.
We demonstrated that our method improved the performance of CTC-based ASR on CSJ and TED-LIUM2 by exploiting the knowledge of BERT.
For future work, we will investigate applying BERT to neural network transducers \cite{Graves12-ST, Zhang20-TT} that are frame-synchronous models and have an autoregressive nature.

\bibliographystyle{IEEEbib}
\bibliography{strings,refs}

\end{document}